# Artistic Style in Robotic Painting; a Machine Learning Approach to Learning Brushstroke from Human Artists

Ardavan Bidgoli, Manuel Ladron De Guevara, Cinnie Hsiung, Jean Oh, Eunsu Kang

*Abstract*— Robotic painting has been a subject of interest among both artists and roboticists since the 1970s. Researchers and interdisciplinary artists have employed various painting techniques and human-robot collaboration models to create visual mediums on canvas. One of the challenges of robotic painting is to apply a desired artistic style to the painting. Style transfer techniques with machine learning models have helped us address this challenge with the visual style of a specific painting. However, other manual elements of style, i.e., painting techniques and brushstrokes of an artist, have not been fully addressed.

We propose a method to integrate an artistic style to the brushstrokes and the painting process through collaboration with a human artist. In this paper, we describe our approach to 1) collect brushstrokes and hand-brush motion samples from an artist, and 2) train a generative model to generate brushstrokes that pertains to the artist's style, and 3) fine tune a stroke-based rendering model to work with our robotic painting setup. We will report on the integration of these three steps in a separate publication. In a preliminary study, 71% of human evaluators find our reconstructed brushstrokes are pertaining to the characteristics of the artist's style. Moreover, 58% of participants could not distinguish a painting made by our method from a visually similar painting created by a human artist.

The code is available at:
https://github.com/Ardibid/ArtisticStyleRoboticPainting

A demo of this project can be found at:
https://youtu.be/UUFIJr9iQuA

## I. INTRODUCTION

During the past few years, a thriving community of interdisciplinary artists and researchers have joined their efforts to explore the boundaries of creative computing and robotics to enter the realm of painting. There is a rich body of literature in robotic painting that addresses various mediums and techniques, spanning from sophisticated line drawing with pen and paper [1] to detailed oil paintings on canvas [2].

One of the challenges of robotic painting is to integrate a distinct artistic style in the outcomes that can be associated with the contributing artist. During the past few years, style-transfer methods have excelled in learning a desired visual style, derived from a source image, and applying it to a target image [3]. This approach has been tested to let robotic paintings mimic the visual styles of well-known artists. However, an artistic style has various aspects. The common style-transfer methods are based on visual features of the source image. Such an approach does not address other essential elements of style that are closely tied to manual skills, for example, painting techniques and unique brushstrokes of an artist.

This paper aims to study the affordances of machine learning generative models to develop a style learner model at the brushstroke level. We hypothesize that training a generative machine learning model on an artist's demonstrations can help us build a model to generate brushstrokes that pertain to the style of an artist. This model can then be used to generate a range of individual brushstrokes to paint an intricate target image. Our approach distances from learning the artistic visual styles; instead, it focuses on the techniques and characteristics of brushstrokes as an intrinsic element of an artistic style.

Our primary contribution is to develop a method to generate brushstrokes that mimic an artist's style. These brushstrokes can be combined with a stroke-based renderer to form a stylizing method for robotic painting processes. This research aims to achieve this goal by developing a learning-based approach to train a model from a collection of an artist's demonstrations as follows:

1- Adapting a stroke-based rendering (SBR) model to convert an image to a series of brush strokes.

2- Training a stylized brushstroke generator based on the collected demonstration by an artist.

3- Feeding the outcomes of the SBR to the stylizer model and execute the strokes with a robotic painting apparatus.

For the SBR model, we utilize the *Learning to Paint* model [4] to render a given image into a sequence of brushstrokes. We modify and retrain the model to match the constraints of our robot platform, an ABB-IRB120 robotic arm. The robotic arm holds a custom-made fixture that could carry a standard acrylic paintbrush and acrylic paint. For the brushstroke generator, we develop 1) a data collection apparatus to collect both brushstrokes, and brush motions, 2) a data processing pipeline to prepare data for the learning process, and a 3) variational autoencoder to learn the style of the brushstrokes and generate new ones.

We evaluate the proposed idea through a set of user studies. In the user study, we investigate three questions:

Ardavan Bidgoli and Manuel Ladron De Guevara are with the School of Architecture, Carnegie Mellon University, Pittsburgh, PA, 15213 USA (814-777-8873; {abidgoli, manuelr}@andrew.cmu.edu).

Cinnie Hsiung and Eunsu Kang are with the School of Computer Science, Carnegie Mellon University, Pittsburgh, PA, 15213 USA (cinnieh@andrew.cmu.edu, kangeunsu@gmail.com).

Jean Oh is with the Robotics Institute, Carnegie Mellon University, Pittsburgh, PA, 15213 USA (jeanoh@cmu.edu).

1- Can participants distinguish a painting made by a robot from a visually similar painting created by a human artist?
2- Can participants distinguish between a brushstroke drawn by an artist from its replay by a robot?
3- Does the brushstroke generator pertain to the characteristics of an artist's brushstrokes?

The remainder of the paper is organized as follows: Section II is dedicated to the background of the robotic painting and the process of integrating style to such paintings. Section III is focused on our approach. We first discuss our robotic painting apparatus; then, we explain our customized SBR model as well as our stylized brushstroke generator. After describing the data collection/preparation process in Section IV, we report on our user studies in Section V and conclude the paper with a discussion.

## II. RELATED WORKS

The applications of computer-controlled machines to help artists in their creative activities—painting in particular—has been a subject of interest since the mid-seventies. Harold Kohen's AARON is a very well-known example in this field [5]. Most recently, the Robot Art Competition [6] has been a cradle for a thriving community of artists and researchers who are interested in robotic painting. The entries to the competition cover a wide range of painting techniques, human-robot interactions, and robotic platforms. We reviewed the projects submitted to this competition between 2016 and 2018 and enriched the pool of samples with a series of recent projects to depict a better image of this field. We reviewed these projects from three points of view: 1) applying an artistic style, 2) image to stroke conversion, and 3) application of machine learning.

### A. Applying artistic style

In the reviewed projects, artists/roboticists interactively engaged in the process of painting [7], posed as the subject of paintings, provided input signals to drive the process [8], [9], let the robot record their painting process and play it back [10], [11], strap their body to the robot to paint on a large canvas [12], or serve as a canvas [13].

However, in these cases, artists rarely served as a source for a distinct artistic style. Among the reviewed projects, only a small number of them, notably Cloud Painter [14], took advantage of machine learning style transfer methods in order to apply the visual style of a source image to the target painted image. Researchers from CMIT-ReART mimicked the artist's style by recording artist's brushstrokes and playing them back using a robot [10].

It is worth mentioning that the artists who entered the realm of performance with robots have directly applied their style to the process. For example, Sougwen Chung's Exquisite Corpus is a performance by the artist and her robots [7], where she applies her style in the process of performing with the robot. None of the projects took advantage of machine learning methods to apply a style to the brushstroke level.

### B. Image to stroke conversion

Converting an image to a series of brushstrokes is a critical task in the robotic painting process. Most of the studied projects relied on image processing algorithms to map the source input image into a series of brushstrokes with limited variations. These brushstrokes range from small dabs [15] to short strokes. Some projects applied advanced techniques such as underpainting to paint an underlying layer of paint on canvas and then switched to the refinement phase to add the details using precise strokes [16].

### C. Machine learning in robotic painting

There is a body of literature dedicated to exploring the affordances of machine learning methods to simulate brush stroke using Generative Adversarial Networks (GANs) [17], and converting images to a series of brushstroke using Reinforcement Learning (RL) [4] with different level of success. However, among the projects that we studied, no one directly applied machine learning to render an image into brushstrokes or simulate the robot painting process.

## III. METHOD

Our approach to applying a style on the brushstroke level to a robotic painting workflow consists of the following parts: a) the hardware (robotic painting apparatus), b) stroke-based rendering to convert a given image to a series of raw brush strokes that can be executed by the robot, and c) a generative model to generate stylized the brushstrokes based on the inputs from an artist. In the following subsections, we describe each component and its implementation details. Table 1 depicts a high-level image of various steps of this paper. We will report on the fourth row in our future publications.

### A. Robotic painting apparatus

We utilize an ABB IRB 120 articulated robotic arm. The work cell consists of a canvas frame, an end-of-the-arm-tool fixture to hold the brush, a set of painting buckets containing five color paints that define the grayscale range, and a cleaning set. In the process of painting, we implemented two separate sets of actions, 1) painting procedure, and 2) auxiliary procedures such as cleaning the brush tip, drying the excessive water, picking designated colors for each stroke, and cleaning off the extra paint from the brush. We generated the controlling codes in RAPID programming language using HAL [18] add-on for Grasshopper visual programming and parametric modeling [19] on Rhinoceros 3d modeling [20] software.

|   | Input | Process on Input | Post Process | Process on raw data | Outcome | Status |
|---|---|---|---|---|---|---|
| 1 | Human Demonstration<br>Sequence of Motions | Data preprocessing<br>Python | Post Process<br>Python/Grasshopper/HAL | Replay on the Robot<br>ABB 120 Robot | Catalogue of Robotic Brush Strokes | Reported Here |
| 2 | Input Image<br>jpg image * | Stroke-Based Render<br>Learning to Paint * | Post Process<br>Grasshopper/HAL | Execute on the Robot<br>ABB 120 Robot | Robotically Painted Portrait | Reported Here |
| 3 | Human Demonstration<br>Brushstrokes as Images | Data preprocessing<br>Python | Train Generative Model<br>VAE | Generating New Brushstrokes<br>Brushstrokes as Images | Catalogue New Brush Strokes | Reported Here |
| 4 | Human Demonstration<br>Sequence of Motions | Data per-processing<br>Python | Train Generative Model<br>VAE | Generating New Brush Motions<br>Sequence of Motions | Catalogue New Brush Motions | Next step |

Table 1. The four steps of study in this research. Rows 1, 2, and 3 are reported in this paper and the 4[th] row is the next step. Items amrked by * are not contributions of this work.

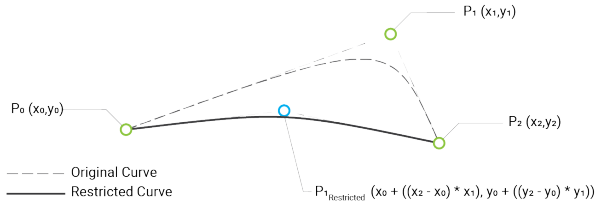

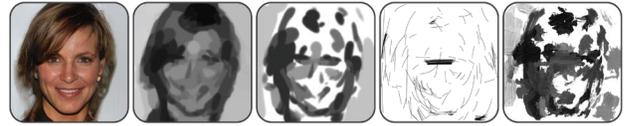

Figure 1. The representation of each stroke as a quadratic Bezier curve: before (dashed line) and after applying the restrictions (continuous line). The restriction effect is reduced for demonstration purposes. The effect of restriction can be seen in Fig. 2 (second image from right).

Figure 2. From left to right: target Image, model-generated stroke set, and final k-means clustered stroke set, paths for robot, painted result. Note the differences between the middle and left image, created by the limitation and affordances of digital-to-physical process.

*B. Stroke-based rendering*

SBR is the process of converting an input image to a series of brushstrokes [21]. We modified the *Learning to Paint* model [4] to generate the sequence of strokes to create the baseline portrait. The model uses reinforcement learning to train the agent to output a set of strokes, based on the current canvas and the target image, to maximize the Wasserstein GAN (WGAN) reward. Similarly to [4], strokes were defined as a list of quadratic Bezier curves (QBC), presented as ($x_0$, $y_0$, $x_1$, $y_1$, $x_2$, $y_2$, $r_0$, $r_1$, $G$), where ($r_0$, $r_1$) are the starting and ending radius of the stroke, and $G$ defines the grayscale color. Constraints are applied on $x_1$ and $y_1$ to prevent the agent from choosing brushstrokes that are too abrupt (Fig. 1).

In contrast to [4], our approach does not allow the agent to modify the transparency of the stroke in order to match the constraints of painting with a real brush and paint. The agent is also restricted to grayscale images rather than the full RGB spectrum to simplify the challenges associated with creating accurate colors. The model is trained on a grayscale converted CelebA dataset [22]. Running the trained model on the *Misun Lean* image [23] generates a sequence of 250 strokes. The final sequence of strokes is generated by using the k-means clustering algorithm on the ($r_0$, $r_1$, $G$) values of the strokes in order to limit the number of colors and the number of brush sizes needed for the robot. The radius of the strokes is limited to up to 25% of the canvas size (Fig. 2). We will discuss this process with more details in section V.A.

*C. A Generative Model to Stylized Brushstroke*

Our efforts on developing a stylized brushstroke generator are focused on two sets of data: 1) brushstrokes, which are the traces of paint on the paper made by the brush and 2) motions, which are the sequences of poses representing the location and orientation of the brush during each brushstroke.

This paper focuses on developing a model that pertains to the artist's style and generate samples of the brushstrokes. We will focus on two other goals in our future work: making a conditional generative model to stylizer model that converts raw strokes to new ones with artist's style, and a generator that generates motions to draw the stylized brush strokes.

Between various available generative models, we decided to work with variational autoencoders (VAE) [24]. The benefits associated with this decision is two-folded; First, it will help us avoiding the GAN's training challenges with respect to the small size of our dataset. Second, while we conducted our survey on the reconstruction aspect of the model, we will heavily investigate its generative capabilities during the next steps of this research. This renders VAE a superior choice compared to AutoEncoders.

Here, the goal is to train a VAE to learn the distribution of an artist's brushstrokes and generate new ones accordingly. To address this goal, we first use the processed data to train a VAE and then draw samples from its latent space. The encoder model is composed of six blocks, each formed by a convolutional 2D, batch normalization, and LeakyReLU layers, followed by a skip capsule. Each skip capsule contains a sequence of convolutional 2D, batch normalization, LeakyReLU, convolution 2D, batch normalization, and a skip connection. The latent space dimension is set to eight (Fig. 3). The decoder network is designed to mirror the encoder. The model is trained for 200 epochs with a batch size of 32, optimized with the Adam optimizer [25] with a learning rate of 0.0005. This model can both collect user samples, map them to the latent space, and then reconstruct them. Also, it can generate new samples through sampling from the latent space.

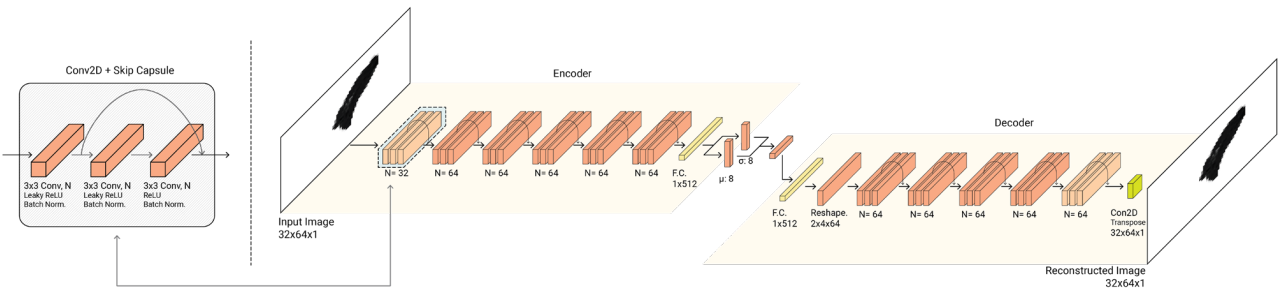

Figure 3. The variational autoEncoder model. The left diagram shows the details of repeated triple-layer blocks in the model.

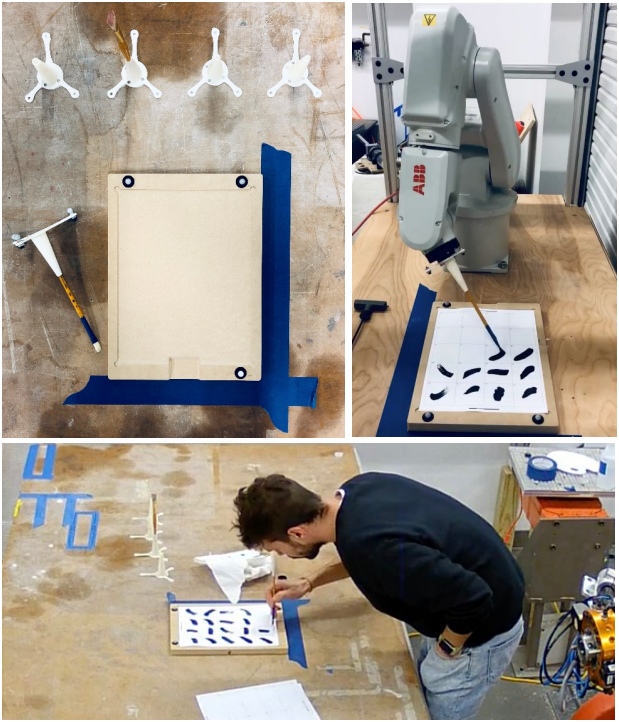

Figure 4. Brush holder and the paper frame for the data collection process (left), demonstration of brushstrokes (bottom), Robotic patining setup (right).

## IV. Data preparation for the learning pipeline

A key idea of our approach is learning from an artist's demonstration. Due to the lack of existing datasets on this end, we design a system to collect a set of data to train our model. In this section, we describe the data collection and data processing steps.

### A. Data collection process

**Hardware**: To record brushstroke motions, we designed, and 3D printed a brush fixture equipped with three reflective markers that form a rigid body for motion capture system. Moreover, to track the position of papers during the data collection sessions, they were fixed in a frame with another set of three reflective markers (Fig. 4, top left). A Motion Capture system with six cameras was used to track these two rigid bodies and reconstruct the brush motions in space with six degrees of freedom.

**Data Recording**: During the data collection process, a user with a background in painting generated over 730 brushstrokes with different lengths, thicknesses, and forms. Brushstrokes were indexed in two types of grid-like datasheets. Each grid contains either 20 2″×2″ square cells or 14 cells combining square cells and 2″×4″ rectangular cells to draw single strokes per cell (Fig. 5).

For each brushstroke sample, we also recorded the corresponding sequence of motions to form our raw data set. Motion capture sessions ran continuously while the user was working on each grid of strokes. The motion capture data include cartesian coordinates and Euler angles of the three markers attached to the brush fixture recorded at 120 frames per second. This continuous stream of raw data is then exported as csv files and post-processed in Grasshopper. The post-processing was focused on isolating each stroke from the continuous stream of motion capture data and matching the brush motion with the corresponding brush stroke on the paper grid.

### B. Data processing

**Brushstrokes**: Each sheet of brushstrokes were scanned and then cropped to isolate each brushstroke as a separated image. While most samples were bounded in a square cell, to have a homogeneous dataset, we saved all brushstrokes as grayscale images with 32×64×1 dimension. Basic image processing operations were applied to correct the white balance, change the image dimensions, and remove the pre-printed cell index from each sample.

**Brush motions**: For the brush motions, we took advantage of the Rhino/Grasshopper 3d environment to visualize the data and inspect the dataset on a page-by-page basis. We used a Z cut-off to break down the continuous stream of data into a series of discrete individual motions for each brushstroke. Using the coordination of the paper holder frame, we centralized each motion around the center of its corresponding grid cell. Low quality and inconsistent samples were discarded at this stage.

To control the variable length for each motion, we chose a fixed length of data as a hyperparameter. Shorter movements were padded to maintain the same length while the longer motions under-sampled to fit in. The array contains six floats, $x$, $y$, $z$ coordinates of the tip of the brush and yaw, pitch, and roll Euler rotations components. Each motion sample forms a 6×60 array. Some of the brushstrokes and motions had to be discarded.

## V. Evaluations

To evaluate our model and verify the research hypotheses, we designed a survey to investigate three questions:
1- Can participants distinguish between a robotically painted image from one made by a robot?
2- Can participants distinguish between a brushstroke drawn by a human artist and its replay by a robot?
3- Do audiences find the generated brush strokes similar to the one that a human user makes?

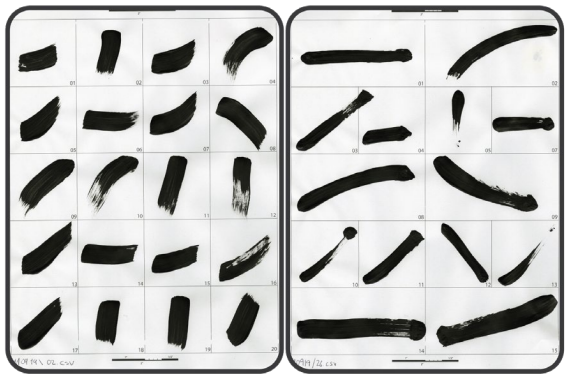

Figure 5. Two grids of samples made by the user. The other grids include a wider range of brushstrokes.

The survey was distributed on Amazon Mechanical Turk (MTurk) as well as students in three universities across the United States. We collected 112 responses, 54 from MTurk, and an additional 58 responses from undergraduate and graduate students.

*A. Survey on robotic painting and human-made painting:*

To investigate the first question, we decided to compare robotically painted portraits with human-drawn ones. On the robotic side, we decided to use one sample made using an SBR (our painting, titled the *Portrait of Misun Lean* (Fig. 6)) and one with a limited swatch of brushstrokes. On the human side, we selected two abstract portraits by contemporary artists and one by Jackson Pollock (Fig. 7).

To create the *Portrait of Misun Lean*, we utilized the optimized *Learning to Paint* model to convert an image of the fictional reporter *Misun Lean* [23] into a series of brushstrokes. The strokes then directly executed using the robotic painting apparatus without applying any artistic style. To do so, we fed the brushstrokes from the *Learning to Paint* model as the input and programmed the robot to follow the provided strokes strictly. For simplicity, we reduced the resolution to 250 brushstrokes. Each stroke had three variables, a) path in the form of a Bezier curve, 2) thickness, limited to four values, and 3) color, limited to a palette of five shades of gray. A Grasshopper definition converted the Bezier curve into a sequence of strictly horizontal target poses. By default, the brush was perpendicular to the Bezier curve to create the thickest brush strokes. For thinner brushstrokes, target planes were rotated on the horizontal plane to compensate for different thicknesses. We utilized HAL add-on to convert these targets into RAPID code, which could run on ABB IRB robotic arms. We also took advantage of ABB drivers to address inverse kinematics and control of the operation. At this point, we did not implement any closed feedback loop.

In the survey, we asked participants, "from the 5 images below, determine which ones are painted by a robotic arm. (You can select more than one. Select 'None' if you think all of them are painted by human artists)". The pool of portraits was composed of digital pictures of 1) *Ghosts of Human-Likeness* by Nicole Coson [26], 2) *Portrait of Misun Lean* by authors, 3) *2Face* by Ryan Hewett [27], 4) *Number 7* by Jackson Pollock [28], and 5. *Untitled 0016* from Artonomous project by Arman Van Pindar [29].

The results demonstrated that among the three human-made options, 45% of participants marked option #1 (Nicole Coson's *Ghosts of Human-Likeness*) and 37% marked #3 (Ryan Hewett's *2Face)* as a robot-made painting, and 33% flagged option #4 (Figure 7).

In comparison, 42% of participants recognized the portrait of *Misun Lean* as a robot-made artifact. The other robotically made option, #5, was flagged only by 34% of participants.

According to the results, more than half of the participants could not distinguish our robotic painting—with its limited number of colors and strokes—from an abstract painting by a human artist that share similar color tones and the level of abstraction. An interesting observation in this question is the number of participants who flagged #5 as the one painted by a robot arm. We assume that this might be a result of a higher number of strokes as well as a wide range of colors.

*B. Can audiences distinguish between a brushstroke drawn by a human artist and its replay by a robot?*

The second question in the survey was aimed to investigate the possibility of using our robotic painting setup to reproduce a series of brushstroke with the same style as an artist.

We selected one of the collected sample grids and processed the motion capture data in Grasshopper/HAL definition to produce the corresponding RAPID program to control the robot with no closed feedback loop. We used the same brush and fixture as the end-of-the-arm-tool on the robot and executed the program. The paint was applied to the brush tip manually as we eliminated all other motions—i.e., refreshing paint—from the raw datasets (Fig. 4, top right).

We aimed to evaluate whether a set of brushstrokes replayed by a robotic arm is differentiable from a set of brushstrokes drawn by a human, or alternatively, do they pertain to the style of the artist? We asked the participants, "Which set of brushstrokes is drawn by a robotic arm? (You can select Both, None, Left, or Right)" and provide them with the replayed brushstrokes as well as the original ones.

Only 40% of participants could select the correct option, while 40% chose the wrong set, 13% said both, and 7% said none (Fig. 8). Thus, we can assume that a well-executed robotic playback can produce brushstrokes that are not distinguishable from the original strokes made by a human artist. In the future, we aim to test the same robotic setup with motions generated by a generative model.

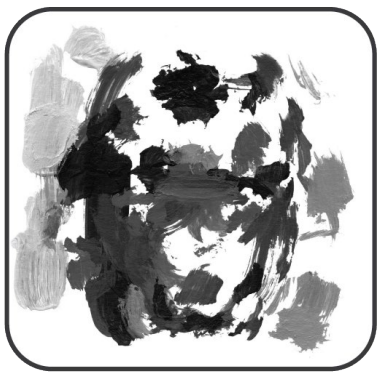

Figure 6. Robotically painted *Portrait of Misun Lean*.

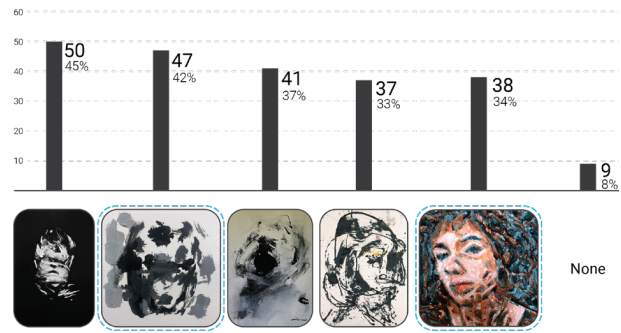

Figure 7. Pool of portraits, from left to right: 1. *Ghosts of Human-Likeness*, 2. *Portrait of Misun Lean*, 3. *2Face*, 4. Number *7* and 5. *Untitled 0016*. Robotic paintings are marked with cyan dashed line.

### C. Do audiences find the generated brush strokes similar to the one that a human user makes?

In the last question of this survey, we aimed to evaluate the quality of brushstrokes that were generated by our generative model. We randomly selected 20 brushstrokes from the dataset and fed them to the model to encode them to the latent space and then decoded them (Fig. 9). We asked participants, *"The set of strokes on the left are drawn by an artist. The set of strokes on the right are recreated by a computer program based on the human-made brushes. To what extent the computer-generated strokes (right) pertain the characteristics of the human-drawn strokes (left)?"*. Participants could select an integer number between 0 and 5, where 0 indicated "No resemblance between the two sets" and 5 indicated that "Styles perfectly match each other." From the 112 participants, over 71% evaluated 3 and higher, with an average of 3.08 and a median of 3 (Fig. 10).

From these results, we can infer that the generator can produce samples that resemble the characteristics of an artist's brushstrokes.

### D. Fidelity and resemblance

In a separate survey, we presented 40 students from the School of Architecture at Carnegie Mellon University with the *Portrait of Misun Lean* and asked them, *"What do you see in the following image? Please respond in less than 5 words."* This question was intended to gauge the level of resolution necessary to achieve an acceptable level of fidelity. The resolution of the outcome is measured by hyperparameters, such as the number of total strokes, colors, and thickness. The *Portrait of Misun Lean* uses 250 strokes, five grayscale colors, and four different thicknesses using only one brush. Based on this survey, the resolution of this setup was not enough to convey the nature of the input, as we saw a variety of answers to this question. The descriptions range from the correct answer *"face"*, to *"a girl with short hair"*, *"chaos"*, *"something I haven't seen before"* or *"a basket of flowers"*.

For the next steps of this research, we aim to increase the resolution by increasing the number of strokes, more granular changes in the stroke thicknesses, and expanding the color spectrum.

## VI. CONCLUSION

We have studied the affordances of Machine Learning Generative models to create a pipeline that could help us integrate an artist painting style at the brushstroke level. Our approach consists of a hardware setup, a customized stroke-based renderer, and a learning pipeline to collect data from an artist, train a model, and generate brushstrokes with the same style as the artist's.

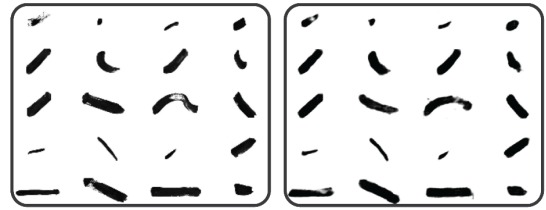

Figure 9. Original brushstrokes (left) and reconstrcuted brushstrokes (right). The model keeps the features of simpler brushstrokes but misses the delicate details of more complex ones.

We create a sample painting using the stroke-based rendering approach and painted it with our robotic apparatus, which more than half of the participants could not recognize as a robotically made artifact. A preliminary study shows a promising result where 71% of participants voted that our reconstructed brushstrokes pertain to the characteristics of the artist's style.

For the next stages of this research, we aim to continue the research in multiple directions:

1- Improving the generative model, we aim to test various state-of-the-art VAEs as well as other generative models such as GANs.

2- Developing a stylizer model conditioned on the SBR outputs, this model can be chained to the SBR and directly generate brushstrokes in the style of the artist.

3- Developing a model to generate a sequence of motions that could result in a given stylized brushstroke. This model can collect stylized brushstrokes from the model mentioned above and create a series of targets for the robotic painting apparatus to paint the brushstrokes on the canvas.

4- Design a pipeline to paint stylized brushstrokes using the robot and enrich the learning dataset with the new samples. We aim to investigate a potential "artist's input vanishing phenomena." If we keep feeding the system with generated motions without mixing them with the original human-generated motions, there would be a point that the human-style would vanish on behalf of a new generated-style. In a cascade of surrogacies, the influence of human agents vanishes gradually, and the affordances of machines may play a more influential role.

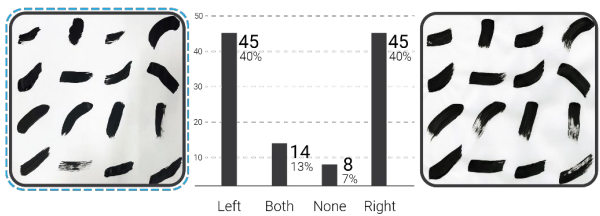

Figure 8. Survey results on distinguishing a robotically replayed brushstroke (left) from the original ones (right). Note that there is no significant difference between the two sets.

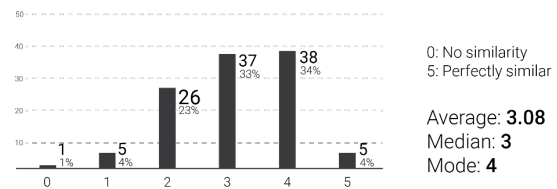

Figure 10. Users' evaluation on the performance of the VAE on reconstructing the original brushstrokes and keeping their visual characteristics.

Under this condition, we are interested in investigating to what extent the human agent's authorship remains in the process.

ACKNOWLEDGMENT

Ardavan Bidgoli and Manuel Ladron De Guevara thank Computational Design Lab (CoDe Lab) and Design Fabrication Lab (DFab) at the School of Architecture, CMU for their generous support and providing the equipment.

The authors deeply appreciate Dr. Barnabas Poczos's help and support. The authors would like to thank Andrew Plesniak for his contribution to the early stages of this research.

The authors would like to express their sincere gratitude towards Nicole Coson, Ryan Hewett, and Pindar Van Arman for granting us permission to use their art pieces in this work.